\documentclass[runningheads]{llncs}

\usepackage{eccv}

\usepackage{eccvabbrv}

\usepackage{amsmath,amssymb}
\usepackage{graphicx}
\usepackage{subcaption}
\usepackage{booktabs}
\usepackage[accsupp]{axessibility}
\usepackage[ruled,vlined]{algorithm2e}
\usepackage[percent]{overpic}

\usepackage{hyperref}
\hypersetup{breaklinks=true,colorlinks=true,citecolor=blue}

\usepackage[section]{placeins}

\graphicspath{
  {last_figure/}
  {last_figure/FigA/}
  {last_figure/FigB/}
  {last_figure/FigC/}
  {last_figure/FigD/}
  {last_figure/FigE/}
  {last_figure/FigF/}
  {last_figure/FigG/}
  {last_figure/FigX/}
  {figures/last_figure/}
  {figures/last_figure/FigA/}
  {figures/last_figure/FigB/}
  {figures/last_figure/FigC/}
  {figures/last_figure/FigD/}
  {figures/last_figure/FigE/}
  {figures/last_figure/FigF/}
  {figures/last_figure/FigG/}
  {figures/last_figure/FigX/}
}

\begin{document}

\title{Attention Frequency Modulation: Training-Free Spectral Modulation of Diffusion Cross-Attention}

\titlerunning{Attention Frequency Modulation (AFM)}







\author{
Seunghun Oh \and
Unsang Park
}

\institute{
Sogang University, Seoul, Republic of Korea\\
\email{\{gnsgus190, uspark\}@sogang.ac.kr}
}

\maketitle

\begin{abstract}
Cross-attention is the primary interface through which text conditions latent diffusion models, yet its step-wise multi-resolution dynamics remain under-characterized, limiting principled training-free control. We cast diffusion cross-attention as a spatiotemporal signal on the latent grid by summarizing token-softmax weights into token-agnostic concentration maps and tracking their radially binned Fourier power over denoising. Across prompts and seeds, encoder cross-attention exhibits a consistent coarse-to-fine spectral progression, yielding a stable time--frequency fingerprint of token competition. Building on this structure, we introduce \emph{Attention Frequency Modulation} (AFM), a plug-and-play inference-time intervention that edits token-wise \emph{pre-softmax} cross-attention logits in the Fourier domain: low- and high-frequency bands are reweighted with a progress-aligned schedule and can be adaptively gated by token-allocation entropy, before the token softmax. AFM provides a continuous handle to bias the spatial scale of token-competition patterns without retraining, prompt editing, or parameter updates. Experiments on Stable Diffusion show that AFM reliably redistributes attention spectra and produces substantial visual edits while largely preserving semantic alignment. Finally, we find that entropy mainly acts as an adaptive gain on the same frequency-based edit rather than an independent control axis.
\end{abstract}

\section{Introduction}
\label{sec:intro}

Diffusion-based text-to-image models achieve state-of-the-art synthesis by progressively denoising text-conditioned latent representations \cite{ho2020ddpm,rombach2022ldm}.
Despite their success, the inference-time internal dynamics that govern how global layout, fine detail, and sample-to-sample variability emerge over denoising steps remain under-explored \cite{ho2022cfg,karras2022edm}.
This gap is not only an interpretability issue but also a controllability bottleneck: widely used controls such as prompt engineering and guidance tuning often act as user-facing heuristics that entangle multiple factors, offering limited insight into what changes inside the model and when those changes occur \cite{ho2022cfg,hertz2022prompt2prompt,chefer2023attendexcite}.

A central yet under-explored component is cross-attention, which injects textual conditioning into the U-Net at multiple resolutions and at every denoising step \cite{rombach2022ldm,vaswani2017attention}.
Because it is repeatedly applied along the diffusion trajectory, cross-attention is a natural carrier of stage-dependent signals (e.g., layout formation early vs.\ detail refinement late) \cite{rombach2022ldm,vaswani2017attention}.
However, existing analyses typically focus on attention heatmaps at a few timesteps or token relevance scores, leaving the time-resolved organization across spatial scales unclear and limiting principled, training-free control.
We ask whether cross-attention exhibits a consistent multi-scale organization over denoising, and whether that structure can be used for controllable, inference-time edits.

\paragraph{Hypothesis and approach.}
We hypothesize that diffusion cross-attention exhibits a robust \emph{coarse-to-fine} spectral progression over denoising, and that steering this progression in logit space provides a training-free knob that \emph{biases} token-competition toward coarser or finer spatial patterns (as measured by our attention-spectrum diagnostics),
rather than directly measuring or guaranteeing image-level layout/detail disentanglement.
To test this hypothesis, we (i) extract a token-agnostic \emph{attention concentration} map from cross-attention and summarize its spectrum over sampling progress, and (ii) intervene by frequency-selective, token-wise reweighting of \emph{pre-softmax} cross-attention logits.
Empirically, encoder cross-attention exhibits the most stable and monotonic spectral trajectory across prompts and seeds, so we use it as the primary locus for both analysis and intervention, while still reporting downstream effects in middle/decoder blocks.

\paragraph{A frequency-domain view.}
We operationalize cross-attention as a spatial signal on the latent grid by mapping each step's token distribution to a token-agnostic concentration statistic (top-$K$) and analyzing its radially binned Fourier power.
This yields a compact time--frequency fingerprint that is stable across prompts and random seeds, revealing a consistent coarse-to-fine evolution inside the denoising trajectory.

\paragraph{Training-free control via AFM.}
Motivated by this fingerprint, we introduce \emph{Attention Frequency Modulation} (AFM), a plug-and-play inference-time intervention that applies frequency-selective reweighting to token-wise \emph{pre-softmax} cross-attention logits.
AFM provides an interpretable control handle by reshaping token competition patterns in logit space over sampling progress, without retraining or architectural modification.
We emphasize that AFM does not ``inject'' high-frequency detail directly.
Instead, it perturbs token competition in logit space, which can \emph{steer the observed} coarse-to-fine spectral progression measured on post-softmax attention summaries.
Unless otherwise stated, we apply AFM to \emph{encoder} cross-attention modules, where the spectral trajectory is most stable, and report downstream effects in middle/decoder blocks.

\begin{figure*}[tb]
\centering
\begin{minipage}{0.99\linewidth}
  \centering
  \includegraphics[width=\linewidth]{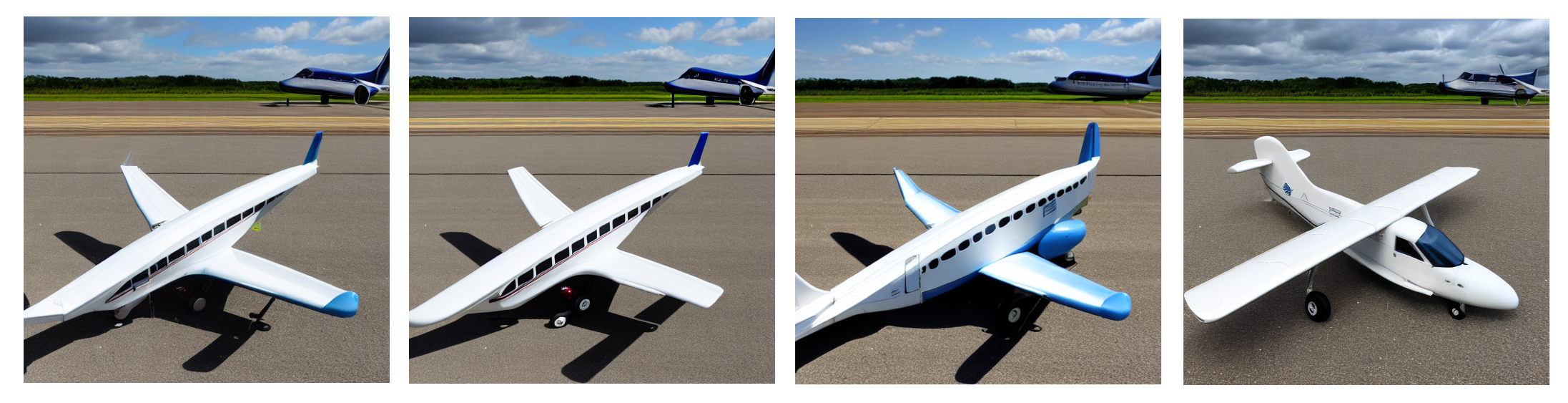}\\
  \small
  \makebox[0.25\linewidth][c]{(a) Baseline}%
  \makebox[0.25\linewidth][c]{(b) SAG}%
  \makebox[0.25\linewidth][c]{(c) FreeU}%
  \makebox[0.25\linewidth][c]{(d) Ours}
\end{minipage}
\caption{Qualitative comparison on Stable Diffusion v1.5 under matched sampling settings (same prompt/seed). (a) Baseline, (b) SAG, (c) FreeU, (d) Ours (AFM).}
\label{fig:teaser}
\end{figure*}

\paragraph{Contributions.}
(i) Frequency characterization of diffusion cross-attention.
We characterize cross-attention using a token-agnostic \emph{concentration} signal
(top-$K$ mean) and track its spectral evolution over denoising,
providing a stable fingerprint of coarse-to-fine \emph{token competition}
across prompts and random seeds.

(ii) Training-Free Attention Frequency Modulation (AFM).
We propose an inference-time method that directly intervenes on cross-attention logits in the frequency domain, applied token-wise before softmax.
AFM-curve suppresses late-stage high-frequency fragmentation in encoder attention
(as measured by the post-softmax top-$K$ spectrum / $\rho_s$) while preserving alignment, indicating controllability over the internal progression.

(iii) Entropy as a secondary gating signal.
We analyze attention entropy as a complementary statistic capturing concentration/dispersion.
When paired with AFM, entropy gates the effective intervention strength and influences variability distributionally.

\paragraph{Related work.}
Training-free diffusion control.
Prior work modifies diffusion inference without retraining, including attention-based editing/excitation and feature-space interventions \cite{hertz2022prompt2prompt,mokady2022nulltext,chefer2023attendexcite,tang2022daam,hong2022sag,si2023freeu,he2024aid_attention_interpolation,lin2024ctrlx_structure_appearance}.
Related controllable generation and parameter-efficient adaptation methods include ControlNet, T2I-Adapter, and LoRA \cite{zhang2023controlnet,mou2023t2iadapter,hu2021lora}.
We include SAG and FreeU as representative training-free baselines for direct comparison in our experiments.
In contrast, we first characterize a stable frequency structure in cross-attention over timesteps and then derive a frequency-aligned, logit-space intervention.
We choose SAG and FreeU as representative training-free interventions that modify the sampling
dynamics without additional training or external models, matching our plug-and-play setting.
Other editing methods that rely on prompt-specific inversion or extra optimization steps are orthogonal
and not directly comparable under our fixed-sampling paired protocol.

Frequency perspectives and mechanistic analyses.
Frequency-domain tools and mechanistic analyses have been used to probe representation bias and diffusion dynamics \cite{rahaman2019spectralbias,tancik2020fourierfeatures,liu2024understanding_cross_self_attn_sd,yi2024understanding_working_mechanism_t2i,jiang2025mechanistic_diffusion_bias}.

Entropy and information measures.
Entropy has long been used to quantify dispersion in probabilistic assignments \cite{shannon1948}.
We show entropy mainly acts as a gating statistic that modulates the effective strength/variability of frequency-based interventions, rather than defining an orthogonal control axis.

\section{Background}
\label{sec:background}

\subsection{Diffusion models and cross-attention along the denoising trajectory}
Diffusion models synthesize images via an iterative denoising process \cite{ho2020ddpm}.
Latent diffusion models (LDMs) perform this process in a learned latent space for efficient high-resolution generation \cite{rombach2022ldm}.
At denoising step $t$, given latent $x_t$ and text condition $c$, the reverse update is
\begin{equation}
x_{t-1} = f_\theta(x_t, t, c),
\end{equation}
where $f_\theta$ is typically a U-Net with multi-resolution blocks: the downsampling path (encoder), the bottleneck (middle), and the upsampling path (decoder).
Text conditioning is injected through cross-attention layers.
With queries $Q$ from latent features and keys/values $K,V$ from text embeddings, a standard cross-attention operation is
\begin{equation}
\mathrm{Attn}(Q,K,V) = \mathrm{softmax}\!\left(\frac{QK^\top}{\sqrt{d}}\right)V.
\end{equation}
Because cross-attention is applied at every denoising step and at multiple spatial resolutions, it naturally admits a stage-dependent view of generation dynamics.
In this work, we treat cross-attention as a temporally evolving internal conditioning signal, rather than a static alignment artifact.
For clarity across schedulers, we later reparameterize the step index by a monotone denoising progress.

\subsection{Cross-attention maps as spatial signals}
For a layer/head at sampling iteration $s$, token-softmax cross-attention yields
$\mathbf{A}_s\in\mathbb{R}^{HW\times T}$ (spatial queries $\times$ tokens).
Since each row sums to 1, naive token summation is uninformative.
We therefore analyze token-agnostic concentration summaries (e.g., mean top-$K$ probability)
as spatial signals on the latent grid and study their spectral evolution over denoising.
All definitions (top-$K$/top-1, entropy statistics) are provided in
Sec.~\ref{sec:method_signal} and Sec.~\ref{sec:method_afm}.

\paragraph{Entropy statistics (used for AFM gating).}
We use the normalized mean token entropy $\bar{\mathcal{H}}^{\mathrm{tok}}_{s}$ (Eq.~\eqref{eq:tok_entropy_mean})
as an optional gate for AFM. Additional entropy diagnostics are deferred to the supplement.

\section{Method}
\label{sec:method}

We present a frequency-domain framework to analyze and edit diffusion cross-attention at inference time.
We first define a measurement pipeline that converts cross-attention into a stable spatial signal and quantifies its spectral evolution over denoising, and then introduce a training-free, plug-and-play intervention that edits token-wise pre-softmax cross-attention logits via frequency-selective reweighting.

\paragraph{Key design choices.}
Two choices are central.
First, we treat cross-attention as a spatiotemporal signal evolving over denoising progress, which motivates a spectral (Fourier) view.
Second, for controllability we operate on token-wise logits (pre-softmax), because token-softmax invariances can render per-query scalar biases broadcast across tokens a no-op (Sec.~\ref{sec:method_afm}).

\subsection{Cross-attention as a spatiotemporal signal}
\label{sec:method_signal}

\paragraph{Cross-attention in latent diffusion U-Nets.}
We consider latent diffusion models (LDMs) with a U-Net backbone and cross-attention conditioning \cite{rombach2022ldm}.
At each denoising step, cross-attention maps latent/image features (queries) to text features (keys/values).
Let a given U-Net block have latent spatial size $H\times W$ (flattened into $HW$ queries) and let the prompt contain $T$ tokens.
For a single attention head, the pre-softmax logits are
\begin{equation}
\mathbf{L}_{s} = \frac{\mathbf{Q}_{s}\mathbf{K}_{s}^{\top}}{\sqrt{d}}
\in \mathbb{R}^{HW \times T},
\end{equation}
and the token-normalized attention weights are
\begin{equation}
\mathbf{A}_{s} = \mathrm{softmax}(\mathbf{L}_{s})
\in \mathbb{R}^{HW \times T},
\end{equation}
where softmax is applied row-wise over the token dimension.
For multi-head attention, we apply the same definitions per head and either analyze heads individually or average derived statistics over heads.

\paragraph{Denoising progress (implementation index).}
We index the $S$ sampling iterations in the exact order executed by the sampler as $s\in\{0,\dots,S{-}1\}$
($s{=}0$: most noisy, $s{=}S{-}1$: most clean), and use the normalized progress $u(s)=s/(S{-}1)$ for all schedules and plots.

\paragraph{Scheduler timesteps (annotation only).}
Some samplers (e.g., DDIM) associate each iteration $s$ with a scheduler timestep $\tau_s$
(e.g., ddim\_timesteps[s]), which is monotone \emph{decreasing} in $s$.
We occasionally use $\tau_s$ only for figure tick labels.
All schedules, interventions, and analyses in this paper are indexed by the sampler step $s$
(or its normalized progress $u(s)$), i.e., the horizontal axis is always early $\rightarrow$ late in $s$.

\paragraph{Why an ``attention map'' is non-trivial under token-softmax.}
A cross-attention row $\mathbf{A}_{s}(i,:)$ is a probability distribution over tokens at each spatial query $i$.
Thus, naive token summation is constant:
$\sum_{j=1}^{T}\mathbf{A}_{s}(i,j)=1$ for all $i$.
This motivates constructing a spatial analysis signal that is (i) defined over $H\times W$, (ii) stable across prompts/seeds, and (iii) non-trivial under token normalization.

\paragraph{Token-agnostic spatial summary (\texttt{top-1} / \texttt{top-$K$}).}
We build a token-agnostic confidence/peakedness map capturing how concentrated
the token distribution is at each location. For each spatial query $i$,
\begin{align}
S_{s}^{\texttt{top}1}(i) &= \max_{j\in\{1,\dots,T\}} \mathbf{A}_{s}(i,j), \\
S_{s}^{\texttt{top}K}(i) &=
\frac{1}{K}\sum_{j\in \mathrm{Top}K(\mathbf{A}_{s}(i,:);K)} \mathbf{A}_{s}(i,j),
\end{align}
and we reshape $S_{s}(\cdot)$ back into $S_{s}\in\mathbb{R}^{H\times W}$.

\paragraph{Why top-$K$ and top-1.}
top-1 can be sensitive to winner-takes-all fluctuations, producing noisier trajectories across seeds.
top-$K$ reduces variance by averaging among the most attended tokens, yielding smoother and more reproducible spectral statistics.
We therefore use top-$K$ in the main analysis and report top-1 as an ablation to show the coarse-to-fine trend is not an artifact of aggregation choice.

\subsection{Frequency decomposition and coarse-to-fine metrics}
\label{sec:method_freq}

\paragraph{2D Fourier transform and normalized power spectrum.}
Given $S_{s}\in\mathbb{R}^{H\times W}$, we compute the 2D Fourier transform $\hat{S}_{s}=\mathcal{F}(S_{s})$.
We define the normalized power spectrum
\begin{equation}
P_{s}(f_x,f_y)=\frac{|\hat{S}_{s}(f_x,f_y)|^2}{\sum_{f_x,f_y}|\hat{S}_{s}(f_x,f_y)|^2},
\qquad \sum_{f_x,f_y}P_{s}(f_x,f_y)=1.
\end{equation}

\paragraph{Radial coordinate and binning (time--frequency matrix).}
Using FFT-shifted coordinates, we define the normalized radius
\begin{equation}
r=\frac{\sqrt{f_x^2+f_y^2}}{r_{\max}}\in[0,1],
\end{equation}
and bin $P_s$ into $B$ radial bins to obtain the radial energy profile
\begin{equation}
E_{s}(b)=\sum_{(f_x,f_y)\in \mathrm{bin}(b)} P_{s}(f_x,f_y),
\qquad \sum_{b=1}^{B}E_{s}(b)=1.
\end{equation}

\paragraph{High-frequency ratio as a coarse-to-fine indicator.}
We summarize coarse-to-fine behavior via the high-frequency (HF) energy ratio using a cutoff radius $r_c$:
\begin{equation}
\rho_{s}=\sum_{b:\, r_b \ge r_c} E_{s}(b).
\label{eq:hf_ratio}
\end{equation}
Since $E_{s}$ is normalized, $\rho_{s}$ measures the fraction of spectral energy in the HF band.
A consistent increase of $\rho_{s}$ over $u(s)$ indicates a progressive shift toward more localized attention structure.

\paragraph{Cutoff choice and robustness.}
The LF/HF split depends on $r_c$, but the trajectory shape can remain robust even when absolute values shift.
We use a default $r_c$ for main plots and test robustness with cutoff sweeps (multiple $r_c$ values), verifying that coarse-to-fine trends and AFM-induced deltas persist.
For intuition, with square grids $r_{\max}\approx\sqrt{0.5^2+0.5^2}$ in cycles/pixel, so $r_c{=}0.25$ corresponds to a radial frequency of about $0.18$ cycles/pixel (roughly a $5\!-\!6$ pixel wavelength).

\paragraph{Intervention diagnostics: deltas and log-ratios.}
To isolate how an inference-time edit changes spectral composition, we compute
\begin{equation}
\Delta\rho_{s} = \rho_{s}^{\text{(target)}} - \rho_{s}^{\text{(ref)}},
\end{equation}
and a frequency-resolved log-ratio
\begin{equation}
R_{s}(b)=\log\frac{E_{s}^{\text{(target)}}(b)+\epsilon}{E_{s}^{\text{(ref)}}(b)+\epsilon},
\end{equation}
with small $\epsilon$ for numerical stability.
$R_{s}(b)$ provides a direct ``what frequencies changed, and when'' explanation.

\subsection{Training-Free Attention Frequency Modulation (AFM)}
\label{sec:method_afm}

We now describe AFM, a plug-and-play inference-time attention editing method that performs token-wise logit-space spectral reweighting during denoising.

\paragraph{Why we edit logits (pre-softmax) instead of attention weights.}
Post-softmax edits of $\mathbf{A}_s$ can be attenuated by the row-wise token normalization.
In particular, token-softmax is invariant to adding a per-query scalar bias broadcast across tokens:
\begin{equation}
\mathrm{softmax}\big(\mathbf{L}_{s}(i,:)+b(i)\mathbf{1}\big)=\mathrm{softmax}\big(\mathbf{L}_{s}(i,:)\big).
\label{eq:shift_invariance}
\end{equation}
Therefore, a purely spatial additive bias shared across tokens cannot change token assignment.
To induce a non-trivial change in $\mathbf{A}_s$, the intervention must be token-dependent and applied before normalization, motivating token-wise edits on $\mathbf{L}_s(:,j)$.

\paragraph{Token-wise spectral reweighting on logit maps.}
For each token $j\in\{1,\dots,T\}$, we reshape the logit column $\mathbf{L}_{s}(:,j)$ into a spatial map
$\mathbf{Z}_{s,j}\in\mathbb{R}^{H\times W}$ and apply FFT:
\begin{equation}
\hat{\mathbf{Z}}_{s,j}=\mathcal{F}(\mathbf{Z}_{s,j}).
\end{equation}
Let $M_{\mathrm{LF}}=\mathbb{I}(r\le r_c)$ and $M_{\mathrm{HF}}=\mathbb{I}(r>r_c)$ with $M_{\mathrm{LF}}+M_{\mathrm{HF}}=\mathbf{1}$.
AFM defines the edited spectrum as
\begin{equation}
\hat{\mathbf{Z}}'_{s,j}
=
\alpha_s^{\mathrm{LF}}\big(\hat{\mathbf{Z}}_{s,j}\odot M_{\mathrm{LF}}\big)
+
\alpha_s^{\mathrm{HF}}\big(\hat{\mathbf{Z}}_{s,j}\odot M_{\mathrm{HF}}\big),
\label{eq:afm_fft}
\end{equation}
followed by $\mathbf{Z}'_{s,j}=\mathcal{F}^{-1}(\hat{\mathbf{Z}}'_{s,j})$ and flattening back to $\mathbf{L}'_{s}(:,j)$.
Finally, we compute the edited attention weights as $\mathbf{A}'_{s}=\mathrm{softmax}(\mathbf{L}'_{s})$.

\paragraph{Hard vs.\ soft masks (ringing reduction).}
Hard binary masks can introduce mild spatial ringing due to sharp spectral boundaries.
Optionally, one may replace $M_{\mathrm{LF}},M_{\mathrm{HF}}$ with a smooth radial transition (e.g., cosine ramp around $r_c$) while preserving interpretability.
Our reported spectral statistics are robust as long as the effective LF/HF separation remains comparable.

\paragraph{Stability details: DC term and real-valuedness.}
Logit maps can contain token-specific global biases that affect overall token competitiveness.
Optionally, we preserve the DC coefficient while applying band scaling to the remaining coefficients.
Radially symmetric masks preserve conjugate symmetry, and the inverse FFT returns a real map up to numerical precision (we take the real part).

\paragraph{Scheduled (curve) scaling aligned with denoising progress (entropy off).}
We use a progress-dependent schedule
\begin{equation}
\alpha^{\mathrm{LF}}_{s}=1+\lambda(1-u(s)),
\qquad
\alpha^{\mathrm{HF}}_{s}=1+\lambda u(s),
\label{eq:curve_schedule}
\end{equation}
where $\lambda$ controls the overall edit strength.

\paragraph{Entropy gating (entropy on).}
We compute mean normalized token entropy from the (unmodified) attention weights
$\mathbf{A}_s=\mathrm{softmax}(\mathbf{L}_s)$:
\begin{equation}
\bar{\mathcal{H}}^{\mathrm{tok}}_{s}
=
\frac{1}{HW\log T}\sum_{i=1}^{HW}
\left(
-\sum_{j=1}^{T} \mathbf{A}_s(i,j)\log(\mathbf{A}_s(i,j)+\epsilon)
\right),
\label{eq:tok_entropy_mean}
\end{equation}
and gate the band scaling as
\begin{equation}
\alpha^{\mathrm{LF}}_{s}=1+\lambda(1-u(s))\,(1+\beta\bar{\mathcal{H}}^{\mathrm{tok}}_{s}),
\qquad
\alpha^{\mathrm{HF}}_{s}=1+\lambda u(s)\,(1+\gamma(1-\bar{\mathcal{H}}^{\mathrm{tok}}_{s})).
\label{eq:entropy_gating}
\end{equation}
When AFM is disabled ($\lambda=0$), entropy gating is a strict no-op by construction.

\paragraph{Important: where the frequency is measured.}
Eq.~\eqref{eq:curve_schedule}--\eqref{eq:entropy_gating} modulate the \emph{logit maps} $\mathbf{L}_s(:,j)$ in the Fourier domain before the token softmax, whereas $\rho_s$ is computed from the \emph{post-softmax} concentration map $S_s^{\texttt{top}K}$.
Because token softmax is nonlinear and induces token competition, the effect of logit-space band scaling on $\rho_s$ is not monotonic; we therefore treat $\rho_s$ as a diagnostic of token-competition patterns, not a direct measure of image-frequency content.

\paragraph{Entropy as an auxiliary gating signal (not an independent edit).}
$\bar{\mathcal{H}}^{\mathrm{tok}}_{s}$ (Eq.~\eqref{eq:tok_entropy_mean}) summarizes token dispersion:
high entropy indicates diffuse token assignment, while low entropy indicates concentrated assignment.
We use it only to gate band scaling (Eq.~\eqref{eq:entropy_gating}), and it becomes a strict no-op when $\lambda=0$.

\paragraph{Layer/block scope.}
AFM is compatible with any subset of U-Net cross-attention modules (\texttt{attn2}).
In our main setting, we apply AFM to \emph{encoder} cross-attention modules only,
and leave self-attention (attn1; context=None) unchanged.
We still log encoder/middle/decoder spectra to separate direct edits from downstream effects.

\begin{algorithm}[t]
\caption{AFM: token-wise logit-space spectral reweighting}
\label{alg:afm}
\KwIn{Cross-attention logits $\mathbf{L}_{s}\in\mathbb{R}^{HW\times T}$ at denoising progress $s$}
\KwOut{Edited logits $\mathbf{L}'_{s}$}
Compute progress $u(s)\in[0,1]$\;
compute $\bar{\mathcal{H}}^{\mathrm{tok}}_{s}$ using Eq.~\eqref{eq:tok_entropy_mean}\;
Set $(\alpha^{\mathrm{LF}}_{s},\alpha^{\mathrm{HF}}_{s})$ using Eq.~\eqref{eq:curve_schedule} (entropy off) or Eq.~\eqref{eq:entropy_gating} (entropy on)\;
\For{$j=1$ \KwTo $T$}{
Reshape $\mathbf{L}_{s}(:,j)$ to $\mathbf{Z}_{s,j}\in\mathbb{R}^{H\times W}$\;
FFT: $\hat{\mathbf{Z}}_{s,j}=\mathcal{F}(\mathbf{Z}_{s,j})$\;
Apply LF/HF reweighting with cutoff $r_c$ (Eq.~\eqref{eq:afm_fft})\;
(Optional) preserve DC coefficient\;
iFFT: $\mathbf{Z}'_{s,j}=\mathcal{F}^{-1}(\hat{\mathbf{Z}}'_{s,j})$\;
Flatten $\mathbf{Z}'_{s,j}$ back into $\mathbf{L}'_{s}(:,j)$\;
}
\Return{$\mathbf{L}'_{s}$}
\end{algorithm}

\section{Experiments}
\label{sec:experiments}

We evaluate Training-Free Attention Frequency Modulation (AFM) along three axes:
(i) attention-level evidence that cross-attention exhibits a consistent coarse-to-fine spectral evolution and that AFM induces controlled spectral redistribution,
(ii) image-level sensitivity under paired generation (LPIPS),
and (iii) text--image alignment stability (CLIP cosine similarity).
Unless otherwise stated, we use paired comparisons with matched prompts and random seeds to isolate the causal effect of AFM.

\subsection{Setup}
\label{sec:exp_setup}

\paragraph{Models.}
Our main attention-spectrum analyses are conducted on Stable Diffusion v1.5 \cite{rombach2022ldm}.
We additionally report image-level robustness (LPIPS/CLIP) on Stable Diffusion v1.4 under the same prompt/seed and sampling protocol.

\paragraph{Sampling (fixed across methods).}
For reproducibility and fair comparison, \emph{within each checkpoint} all methods use the same scheduler,
number of steps, CFG scale, resolution, and negative prompt.
We generate $512{\times}512$ images using a DDIM scheduler with $S{=}50$ denoising steps and guidance scale $7.5$,
and use paired sampling with fixed random seeds $\{2025,2026,2027,2028\}$ for every prompt.
Only the inference-time intervention (Baseline / SAG / FreeU / AFM) differs.

\paragraph{Prompt sets.}
We use two prompt sources:
(i) a COCO-2017 validation caption subset \cite{lin2014coco} ($N_{\text{prompt}}{=}50$),
and (ii) a LAION caption subset (caption-like samples) \cite{schuhmann2022laion5b} ($N_{\text{prompt}}{=}100$).
We use $N_{\text{seed}}{=}4$ fixed seeds per prompt, yielding $N_{\text{pair}}=N_{\text{prompt}}\times N_{\text{seed}}$ paired generations (COCO: 200; LAION: 400).

\paragraph{AFM configuration.}
Unless specified otherwise, AFM uses top-$K$ aggregation with $K{=}8$ for attention analysis
and a radial cutoff $r_c{=}0.25$ for LF/HF separation.
We apply logit-space spectral reweighting to \emph{encoder} cross-attention modules only,
and log spectra for encoder/middle/decoder to measure both direct and downstream effects on attention dynamics.
We use the curve schedule in Eq.~\eqref{eq:curve_schedule} with $\lambda=0.2$.
For entropy gating (when enabled), we use $(\beta,\gamma)=(20,4)$ in Eq.~\eqref{eq:entropy_gating}.

\paragraph{Intervention scope vs.\ logging scope.}
Unless otherwise stated, AFM is applied \emph{only} to encoder cross-attention modules (attn2 in the downsampling path / input\_blocks).
Middle and decoder cross-attention modules are not directly modified; we report their spectra only as \emph{downstream diagnostics} induced by the altered denoising trajectory.

\paragraph{Compared settings.}
We evaluate:
Baseline (AFM disabled),
curve (AFM-curve; entropy gating off),
and curve + entropy (AFM-curve; entropy gating on).
We also include a negative control Baseline + entropy, where entropy is computed/enabled but AFM strength is set to $\lambda{=}0$; by construction, this should produce identical outputs to Baseline.

\paragraph{Attention logging and aggregation.}
We instrument cross-attention modules to record step-wise statistics.
For frequency analysis, we report encoder cross-attention (6 layers averaged), which yields the most stable trajectories.
We convert token-softmax attention weights into a token-agnostic spatial signal using top-$K$ aggregation (Sec.~\ref{sec:method_signal}), then compute frequency statistics per denoising step.

\paragraph{Step convention.}
All plots use denoising progress $u(s)$ (early $\rightarrow$ late); tick labels optionally show the corresponding scheduler timesteps $\tau_s$ (Sec.~\ref{sec:method_signal}).

\paragraph{Metrics.}
Attention-level: time--frequency heatmaps (radially binned normalized FFT power over steps),
HF energy ratio $\rho_s$ (Eq.~\ref{eq:hf_ratio}),
and log-ratio heatmaps between settings.
Image-level: LPIPS \cite{zhang2018lpips} on paired outputs.
Alignment: CLIP cosine similarity distributions (ViT-B/32) \cite{hessel2021clipscore,radford2021clip}.
Additionally, we report band-wise LPIPS (LPIPS$_{\mathrm{low}}$/LPIPS$_{\mathrm{high}}$) by decomposing outputs into Gaussian low-pass and residual high-pass components to quantify structure vs.\ detail changes.

\subsection{Attention-level results}
\label{sec:exp_results}

\begin{figure*}[tb]
\centering
\includegraphics[width=0.94\linewidth]{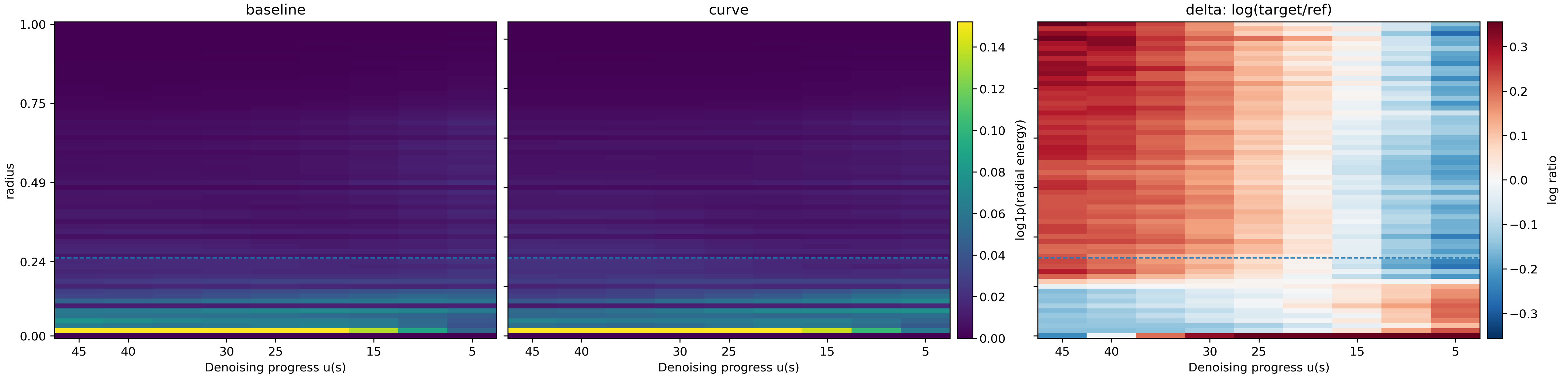}
\caption{Time--frequency evolution of encoder cross-attention (top-$K$, mean).
Left/middle: normalized radial energy distributions for Baseline and AFM-curve.
Right: log energy ratio $\log(E_{\text{curve}}/E_{\text{baseline}})$, highlighting frequency bands amplified/suppressed by AFM over denoising progress.
\textbf{The x-axis is denoising progress $u(s)$ (early $\rightarrow$ late); tick labels show the corresponding DDIM scheduler timesteps $\tau_s$ (decreasing).}
Dashed line indicates the HF cutoff radius $r_c$ used in $\rho_s$.}
\label{fig:timefreq}
\end{figure*}

\paragraph{Baseline cross-attention exhibits a coarse-to-fine spectral evolution.}
We first establish a step-wise spectral signature of diffusion conditioning dynamics.
Fig.~\ref{fig:timefreq} shows the time--frequency evolution of encoder cross-attention under Baseline.
Energy concentrates near low radius early and progressively shifts outward as denoising proceeds, consistent with a coarse-to-fine transition in attention structure.
We observe a consistent coarse-to-fine progression throughout denoising, supporting that the spectral dynamics are stable across runs.

\begin{figure*}[t]
\centering
\begin{subfigure}{0.49\textwidth}
  \centering
  \includegraphics[width=\linewidth]{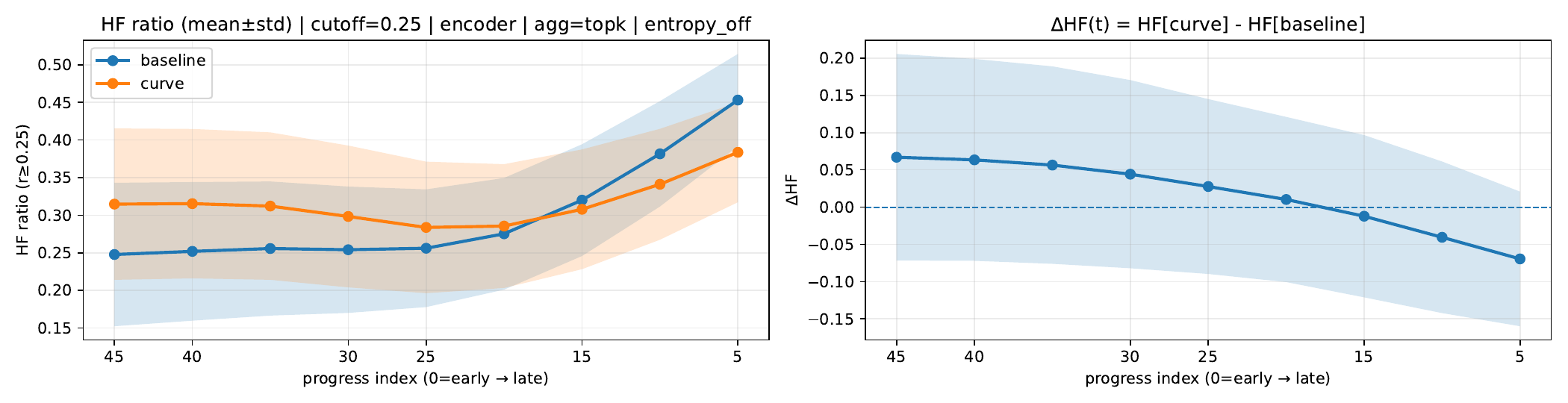}
  \caption{HF ratio $\rho_s$ and $\Delta\rho_s$.}
\end{subfigure}\hfill
\begin{subfigure}{0.49\textwidth}
  \centering
  \includegraphics[width=\linewidth]{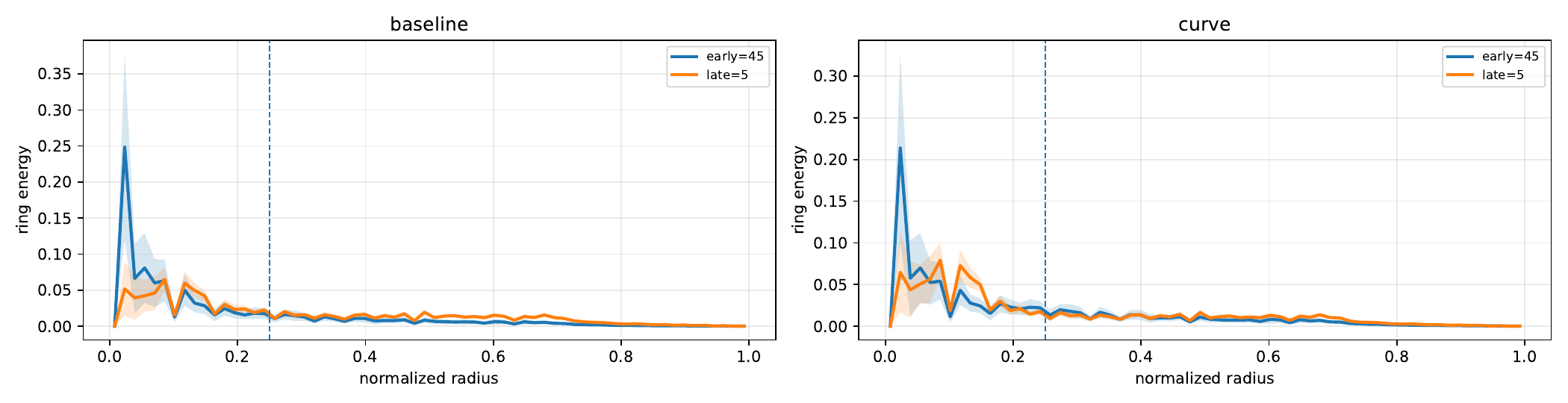}
  \caption{Radial energy profiles.}
\end{subfigure}
\caption{Quantitative summary of coarse-to-fine and AFM effects (encoder, top-$K$).
\textbf{(a) uses denoising progress $u(s)$ (early $\rightarrow$ late); tick labels show decreasing DDIM timesteps $\tau_s$. (b) shows radial energy profiles over normalized radius $r$.}}
\label{fig:freq_summary}
\end{figure*}

\paragraph{AFM induces time-aligned spectral redistribution.}
AFM-curve is designed to modulate spectral allocation over denoising.
In practice, encoder attention shows a consistent reduction in the late-stage HF ratio $\rho_s$ of the post-softmax top-$K$ concentration map (our diagnostic) under AFM-curve, indicating suppression of high-frequency fragmentation relative to the natural coarse-to-fine progression. Note that this ``HF suppression'' refers to $\rho_s$ measured on the post-softmax top-$K$ attention summary $S_s$,
not the raw logit spectra being scaled by Eq.~\eqref{eq:curve_schedule}.
Fig.~\ref{fig:freq_summary} summarizes this effect using the HF ratio $\rho_s$ and $\Delta\rho_s$ (curve minus Baseline).

\paragraph{Late-stage summary.}
We report the HF ratio averaged over the last 20\% of steps (10 steps for $S{=}50$) and its paired difference $\Delta\rho_{\text{late}}$ (curve minus Baseline).

\paragraph{Statistical significance of attention-spectrum changes.}
We evaluate the paired effect of AFM-curve on the encoder HF ratio $\rho_s$ (Eq.~\ref{eq:hf_ratio}) using matched prompt/seed generations.
Bootstrap confidence intervals on the mean $\Delta\rho_{\text{late}}$ exclude zero, and the direction is consistent across pairs.
This indicates that AFM-curve suppresses late-stage high-frequency fragmentation in encoder cross-attention, even though it induces substantial perceptual changes at the image level (LPIPS) with largely overlapping CLIP cosine similarity distributions.
We repeat the analysis with alternative spatial summaries (e.g., top-1 and varying top-$K$) and observe the same direction of $\Delta\rho_{\text{late}}$, suggesting the trend is not a top-$K$ artifact.

\paragraph{Robustness across $r_c$ and sample size.}
To verify that the observed encoder HF-ratio shift is not an artifact of a particular
\emph{metric} cutoff in $\rho_s$ (Eq.~\ref{eq:hf_ratio}) or a small sample size,
we sweep the HF cutoff used in the metric, $r_c \in \{0.20,0.25,0.30\}$,
and subsampled prompt--seed pairs $N_{\text{sub}}\in\{50,100\}$ (top-$K{=}8$).
We define
$\Delta\rho_{\text{late}}=\frac{1}{|\mathcal{S}_{\text{late}}|}\sum_{s\in\mathcal{S}_{\text{late}}}
(\rho_s^{\text{curve}}-\rho_s^{\text{baseline}})$
with $\mathcal{S}_{\text{late}}=\{s:\,u(s)\ge 0.8\}$.
Across all sweeps, the encoder shows a highly consistent negative shift
($\Delta\rho_{\text{late}}<0$ for 98--99\% of pairs), with mean $\Delta\rho_{\text{late}}$
in the range $[-0.063,\,-0.040]$ (Tab.~\ref{tab:delta_late_sweep}).
Middle/decoder show negative means but reduced sign consistency
(middle: 72--78\% negative; decoder: 56--64\% negative), indicating a weaker and more variable
downstream tendency.

\begin{table}[t]
\centering
\scriptsize
\setlength{\tabcolsep}{4pt}
\caption{Robustness of late-stage HF-ratio shift $\Delta\rho_{\mathrm{late}}$ across
$r_c \in \{0.20, 0.25, 0.30\}$ and $N_{\mathrm{sub}} \in \{50,100\}$ (top-$K{=}8$).
$\Delta\rho_{\mathrm{late}}$ is averaged over the last 20\% denoising steps ($u\ge 0.8$).
AFM is applied only to the encoder cross-attention; middle/decoder reflect downstream changes.
We report the fraction of pairs with $\Delta\rho_{\mathrm{late}}<0$ and the min--max range of the mean $\Delta\rho_{\mathrm{late}}$ across sweeps.}
\label{tab:delta_late_sweep}
\begin{tabular}{lcc}
\toprule
Stage & Neg. pairs (\%) & Mean $\Delta\rho_{\mathrm{late}}$ (min--max) \\
\midrule
Encoder & 98--99 & [$-0.063,\ -0.040$] \\
Middle  & 72--78 & [$-0.0346,\ -0.0317$] \\
Decoder & 56--64 & [$-0.0161,\ -0.0092$] \\
\bottomrule
\end{tabular}
\end{table}


\paragraph{Block-wise diagnostics across U-Net stages.}
All U-Net stages exhibit non-trivial spectral structure over denoising, but encoder cross-attention shows the most stable and monotonic coarse-to-fine trajectory across prompts and seeds.
Accordingly, we apply AFM only to encoder cross-attention and treat middle/decoder cross-attention as downstream diagnostics.
Despite modifying only the encoder, the mean late-stage HF-ratio shift is negative in middle and decoder, but with reduced sign consistency (Tab.~\ref{tab:delta_late_sweep}), indicating weaker and more variable downstream effects.

\subsection{Image-level controllability and alignment}
\label{sec:exp_image}

We compare AFM against SAG and FreeU as representative training-free inference-time baselines under identical prompts, seeds, and sampling settings \cite{hong2022sag,si2023freeu}.
Prompts are sampled from COCO and LAION, and results are averaged over fixed seeds (2025--2028).
All methods share the same scheduler, step count, CFG scale, and output resolution to isolate the effect of the intervention.

\paragraph{Baselines and our setting.}
For SAG we use sag\_scale=1.0.
For FreeU we use $(b_1,b_2,s_1,s_2)=(1.1,1.2,0.9,0.2)$.
For AFM-curve (ours) we apply logit-space frequency modulation to cross-attention with
$r_c=0.25$ and the timestep-scheduled curve in Eq.~\eqref{eq:curve_schedule}, unless stated otherwise.
We use default/recommended hyperparameters from the authors' official implementations, without additional tuning. We evaluate all methods on both SD v1.5 (main) and SD v1.4 (checkpoint robustness) for image-level metrics under the same prompt/seed and sampling protocol.

\paragraph{Paired perceptual deviation (LPIPS).}

We quantify image-level sensitivity using LPIPS on paired prompt/seed generations (Tab.~\ref{tab:lpips_ext}).

AFM induces substantial perceptual deviations relative to Baseline.
As a negative control, enabling entropy while disabling AFM ($\lambda=0$) yields identical outputs (LPIPS$=0$), confirming that entropy computation is a strict no-op and only modulates the effective strength when paired with AFM.

\begin{table*}[t]
\centering
\scriptsize
\setlength{\tabcolsep}{3pt}
\caption{\textbf{Paired LPIPS$\uparrow$ (mean$\pm$std) under matched prompt/seed sampling.} Higher indicates larger perceptual deviation from Baseline. Results are reported for SD v1.5 (main) and SD v1.4 (checkpoint robustness).}
\label{tab:lpips_ext}
\begin{tabular}{lcccc}
\toprule
& \multicolumn{2}{c}{SD v1.5} & \multicolumn{2}{c}{SD v1.4} \\
\cmidrule(lr){2-3}\cmidrule(lr){4-5}
Comparison & COCO & LAION & COCO & LAION \\
\midrule
Baseline vs.\ AFM-curve              & 0.237$\pm$0.138 & 0.249$\pm$0.142 & 0.232$\pm$0.132 & 0.258$\pm$0.144 \\
Baseline vs.\ AFM-curve + entropy    & \textbf{0.409$\pm$0.131} & \textbf{0.419$\pm$0.149} & \textbf{0.405$\pm$0.134} & \textbf{0.417$\pm$0.143} \\
Baseline vs.\ Baseline (entropy, $\lambda{=}0$) & 0.000$\pm$0.000 & 0.000$\pm$0.000 & 0.000$\pm$0.000 & 0.000$\pm$0.000 \\
\midrule
Baseline vs.\ FreeU               & \underline{0.300$\pm$0.131} & \underline{0.302$\pm$0.158} & \underline{0.296$\pm$0.117} & \underline{0.294$\pm$0.112} \\
Baseline vs.\ SAG                 & 0.111$\pm$0.084 & 0.125$\pm$0.108 & 0.106$\pm$0.055 & 0.121$\pm$0.071 \\
\bottomrule
\end{tabular}
\end{table*}

\begin{table}[t]
\centering
\scriptsize
\setlength{\tabcolsep}{3pt}
\caption{\textbf{Band-wise LPIPS decomposition (COCO; $N{=}200$).}
We decompose each output image into $I_{\mathrm{low}}=\mathcal{G}_{\sigma}(I)$ (Gaussian blur; $\sigma{=}4$ at $512{\times}512$) and
$I_{\mathrm{high}}=\mathrm{clip}_{[0,1]}(I-I_{\mathrm{low}}+0.5)$, and compute LPIPS on each band after resizing to $256{\times}256$.
LPIPS$_{\mathrm{low}}$ proxies structure/layout change and LPIPS$_{\mathrm{high}}$ proxies detail/texture change.
High/Low is the per-pair ratio LPIPS$_{\mathrm{high}}$/LPIPS$_{\mathrm{low}}$ averaged over pairs (undefined for the exact no-op).
$P(\mathrm{high}>\mathrm{low})$ is the fraction of pairs where LPIPS$_{\mathrm{high}}$ exceeds LPIPS$_{\mathrm{low}}$.}
\label{tab:band_lpips}
\begin{tabular}{lcccc}
\toprule
Setting & LPIPS$_{\mathrm{low}}$ & LPIPS$_{\mathrm{high}}$ & High/Low & $P(\mathrm{high}>\mathrm{low})$ \\
\midrule
AFM-curve & 0.171$\pm$0.115 & 0.188$\pm$0.112 & 1.21$\pm$0.28 & 78.0\% \\
AFM-curve + entropy & 0.312$\pm$0.116 & 0.329$\pm$0.110 & 1.08$\pm$0.15 & 72.5\% \\
Entropy only ($\lambda{=}0$) & 0.000$\pm$0.000 & 0.000$\pm$0.000 & -- & -- \\
\bottomrule
\end{tabular}
\end{table}

\paragraph{Quantifying controllability: structure vs.\ detail (band-wise LPIPS).}
To make the notion of ``control'' explicit at the image level without ground-truth layouts, we decompose each generated image $I$ into a low-frequency component $I_{\mathrm{low}}=\mathcal{G}_{\sigma}(I)$ (Gaussian blur) and a high-frequency residual $I_{\mathrm{high}}=\mathrm{clip}(I-I_{\mathrm{low}}+0.5)$.
We then compute LPIPS on each band between paired baseline and edited outputs.
LPIPS$_{\mathrm{low}}$ serves as a proxy for coarse structure/layout changes, while LPIPS$_{\mathrm{high}}$ captures fine-detail/texture changes.

As shown in Tab.~\ref{tab:band_lpips}, AFM tends to induce larger perceptual changes in the high-frequency residual than in the low-frequency component
(\,LPIPS$_{\mathrm{high}}$ > LPIPS$_{\mathrm{low}}$ for the majority of pairs; see $P(\mathrm{high}>\mathrm{low})$\,),
suggesting that AFM steers generation changes toward fine-detail/texture variations more than coarse structure.

\paragraph{Entropy acts as gain control for frequency-based editing.}

Entropy gating amplifies the frequency-based edit (Tab.~\ref{tab:lpips_ext}); enabling entropy with AFM disabled ($\lambda=0$) is a strict no-op.

\paragraph{Text--image alignment (CLIP cosine similarity).}
We evaluate prompt alignment using CLIP cosine similarity distributions (Tab.~\ref{tab:clip_ext}).
Distributions largely overlap across settings with small mean differences, suggesting AFM redistributes how conditioning manifests rather than collapsing it.
Baseline comparisons are summarized in Tab.~\ref{tab:clip_ext}.
The same qualitative trend is observed on SD v1.4, indicating checkpoint-level robustness.

\begin{table*}[t]
\centering
\scriptsize
\setlength{\tabcolsep}{2pt}
\caption{\textbf{CLIP cosine similarity$\uparrow$ (ViT-B/32; mean$\pm$std) under matched prompt/seed sampling.} Higher indicates better text--image alignment. Results are reported for SD v1.5 (main) and SD v1.4 (checkpoint robustness).}
\label{tab:clip_ext}
\begin{tabular}{lcccc}
\toprule
& \multicolumn{2}{c}{SD v1.5} & \multicolumn{2}{c}{SD v1.4} \\
\cmidrule(lr){2-3}\cmidrule(lr){4-5}
Setting & COCO & LAION & COCO & LAION \\
\midrule
Baseline                & \underline{0.318$\pm$0.031} & 0.306$\pm$0.038 & \textbf{0.320$\pm$0.027} & 0.306$\pm$0.039 \\
AFM-curve               & \textbf{0.319$\pm$0.030} & 0.306$\pm$0.039 & \underline{0.318$\pm$0.029} & 0.304$\pm$0.040 \\
AFM-curve + entropy     & 0.316$\pm$0.030 & 0.303$\pm$0.040 & 0.317$\pm$0.030 & 0.303$\pm$0.038 \\
\midrule
FreeU & 0.307$\pm$0.031 & \textbf{0.310$\pm$0.037} & 0.307$\pm$0.029 & \textbf{0.310$\pm$0.034} \\
SAG   & 0.305$\pm$0.032 & \underline{0.307$\pm$0.036} & 0.304$\pm$0.029 & \underline{0.308$\pm$0.033} \\
\bottomrule
\end{tabular}
\end{table*}

\section{Conclusion}
We presented a frequency-domain view of diffusion cross-attention by interpreting attention-derived concentration maps as spatial signals on the latent grid, revealing a stable coarse-to-fine spectral progression over denoising. Building on this structure, we introduced \emph{Attention Frequency Modulation} (AFM), a training-free inference-time intervention that edits token-wise \emph{pre-softmax} cross-attention logits in the Fourier domain with a progress-aligned low/high-frequency schedule. AFM provides an interpretable inference-time knob to \emph{bias} how conditioning manifests across spatial scales in attention-derived diagnostics, which in turn tends to yield stronger changes in image high-frequency residuals under our proxy evaluation. We further find that attention entropy mainly acts as an adaptive gain for the same frequency-based edit and is a strict no-op when AFM strength is zero.

\noindent\textbf{Limitations and potential negative impact.}
Our coarse-to-fine interpretation is based on spatial-frequency statistics of attention-derived top-$K$ concentration maps, which quantify token-competition structure on the latent grid. These signals are \emph{proxies} and do not directly measure image Fourier content, object layout, or semantic detail.
Accordingly, our image-level evaluation (band-wise LPIPS) should be interpreted as heuristic evidence rather than a ground-truth disentanglement metric.


\FloatBarrier

\bibliographystyle{splncs04}
\bibliography{main}

\end{document}